\documentclass[letterpaper, 10 pt, conference]{ieeeconf}  
\usepackage{graphicx}
\usepackage{subcaption}
\usepackage{latexsym}
\usepackage{amssymb}
\usepackage{amsmath}
\usepackage{booktabs}
\usepackage{color}
\usepackage{bm}
\usepackage{algorithm}
\usepackage{algorithmic}
\newtheorem{theorem}{Theorem}
\newtheorem{lemma}[theorem]{Lemma}

\newtheorem{proposition}[theorem]{Proposition}

\newtheorem{definition}{Definition}
\IEEEoverridecommandlockouts                              

\title{Pareto Front Shape-Agnostic Pareto Set Learning in Multi-Objective Optimization
}

\author{Rongguang Ye$^{1}$, Longcan Chen$^{1}$, Wei-Bin Kou$^{2}$, Jinyuan Zhang$^{1}$ and Hisao Ishibuchi$^{1}$
\thanks{$^{1}$Department of Computer Science and Engineering, Southern University of Science and Technology, Shenzhen 518055, China.
        { \{yerg2023, 12250061\}@mail.sustech.edu.cn,}{\{zhangjy, hisao\}@sustech.edu.cn}. Jinyuan Zhang and Hisao Ishibuchi are corresponding authors.}
\thanks{$^{2}$Department of Electrical and Electronic Engineering, The University of
Hong Kong, Hong Kong 999077, China.  {wbkou@connect.hku.hk}}%
}

\begin{document}

\maketitle
\thispagestyle{empty}
\pagestyle{empty}

\begin{abstract}

Pareto set learning (PSL) is an emerging approach for acquiring the complete Pareto set of a multi-objective optimization problem. Existing methods primarily rely on the mapping of preference vectors in the objective space to Pareto optimal solutions in the decision space. However, the sampling of preference vectors theoretically requires prior knowledge of the Pareto front shape to ensure high performance of the PSL methods. Designing a sampling strategy of preference vectors is difficult since the Pareto front shape cannot be known in advance. To make Pareto set learning work effectively in any Pareto front shape, we propose a Pareto front shape-a\underline{g}nostic \underline{P}areto \underline{S}et \underline{L}earning (GPSL) that does not require the prior information about the Pareto front. The fundamental concept behind GPSL is to treat the learning of the Pareto set as a distribution transformation problem. Specifically, GPSL can transform an arbitrary distribution into the Pareto set distribution. We demonstrate that training a neural network by maximizing \textit{hypervolume} enables the process of distribution transformation. Our proposed method can handle any shape of the Pareto front and learn the Pareto set without requiring prior knowledge. Experimental results show the high performance of our proposed method on diverse test problems compared with recent Pareto set learning algorithms.

\end{abstract}

\section{Introduction}
 Multi-objective optimization problems (MOPs) are frequently faced in the real world \cite{tanabe2020easy,liao2008multiobjective,ashby2000multi}. These problems require optimizing multiple conflicting objectives at the same time. Notably, no single solution can optimize all objectives simultaneously in MOPs. Instead, there are optimal solutions (commonly called Pareto optimal solutions) with different trade-offs among involved objectives. MOPs aim to find all Pareto optimal solutions (commonly known as the Pareto set) \cite{ehrgott2005multicriteria,miettinen1999nonlinear}. The image of the Pareto set in the objective space is the Pareto front. In MOPs, multi-objective evolutionary algorithms (MOEAs) stand out as widely employed methods \cite{zitzler1999evolutionary,zhou2011multiobjective}. These algorithms iteratively optimize a population of solutions and finally obtain a solution set that is well distributed on the Pareto front. In continuous MOPs with $m$ objectives, the shapes of the Pareto set and Pareto front are usually a $(m-1)$-dimensional manifold \cite{hillermeier2001generalized,zhang2008rm}. Especially, for many-objective optimization problems ($m\geq 4$), it is difficult to approximate the whole Pareto front (and Pareto set) using a limited number of solutions obtained by an MOEA. Going beyond existing MOEAs that have deficient capabilities in the continuous manifold, neural networks emerge as a powerful method to obtain the entire Pareto set.
 
\begin{figure}
    \centering
    \includegraphics[width=0.8\linewidth]{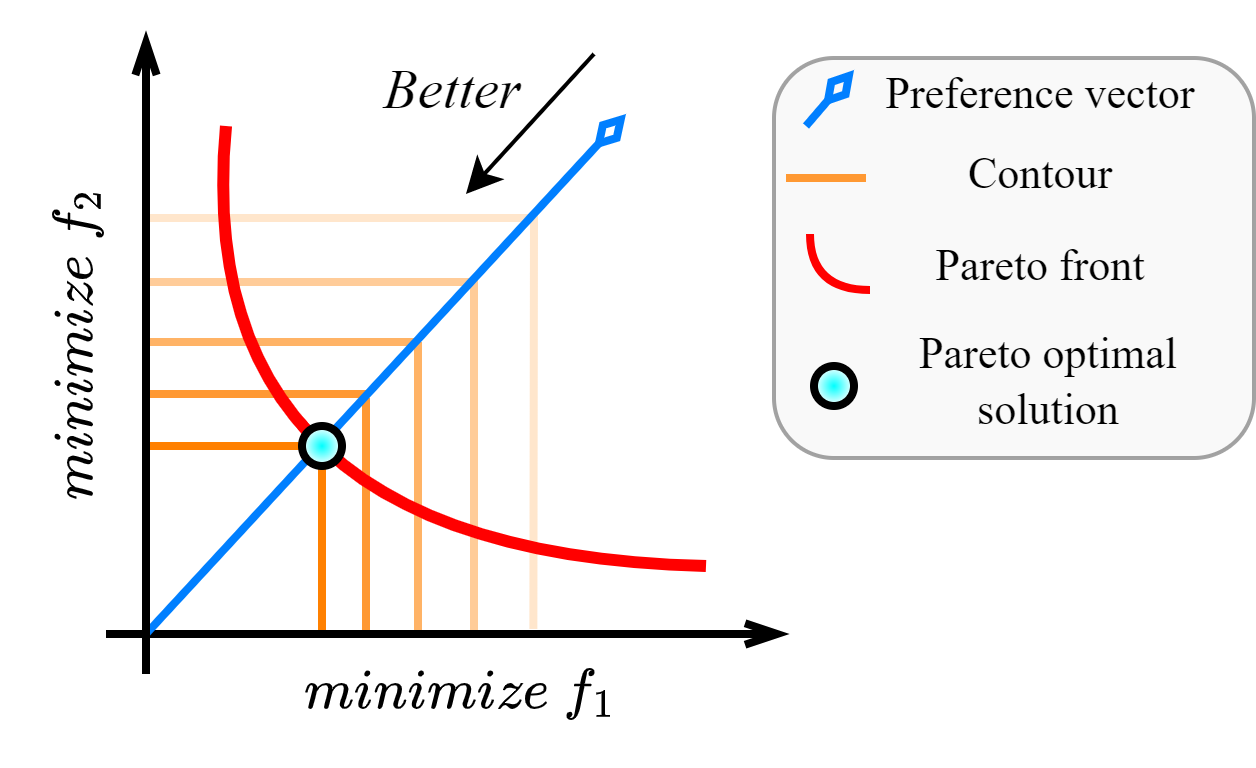}
    
    \caption{\textbf{Overview of Preference-based PSL methods.} Orange lines are contours of weighted Techebycheff loss function (in Section \ref{PPSL}). The optimal solution obtained by Preference-based PSL methods is the intersecting point of Pareto front and given preference vector.}
    \label{fig:tch_goal}
    \vspace{-0.4cm}
\end{figure}

\begin{figure*}[htbp]
    \centering
    \begin{subfigure}[b]{0.22\textwidth}
        \centering
        \includegraphics[width=\linewidth]{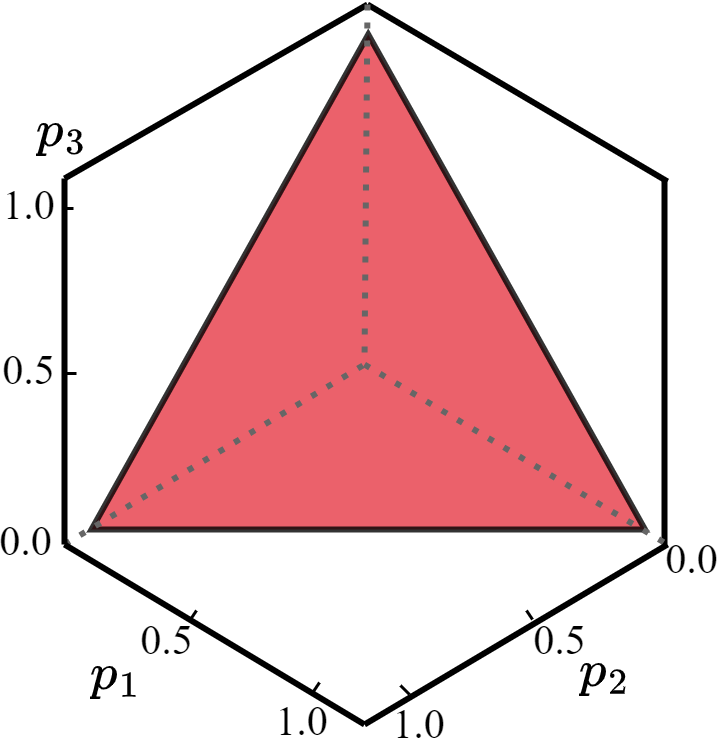}
        \caption{3D triangular preference sampling space}
        \label{ma}
    \end{subfigure}
    \hfill
    \begin{subfigure}[b]{0.22\textwidth}
        \centering
        \includegraphics[width=\linewidth]{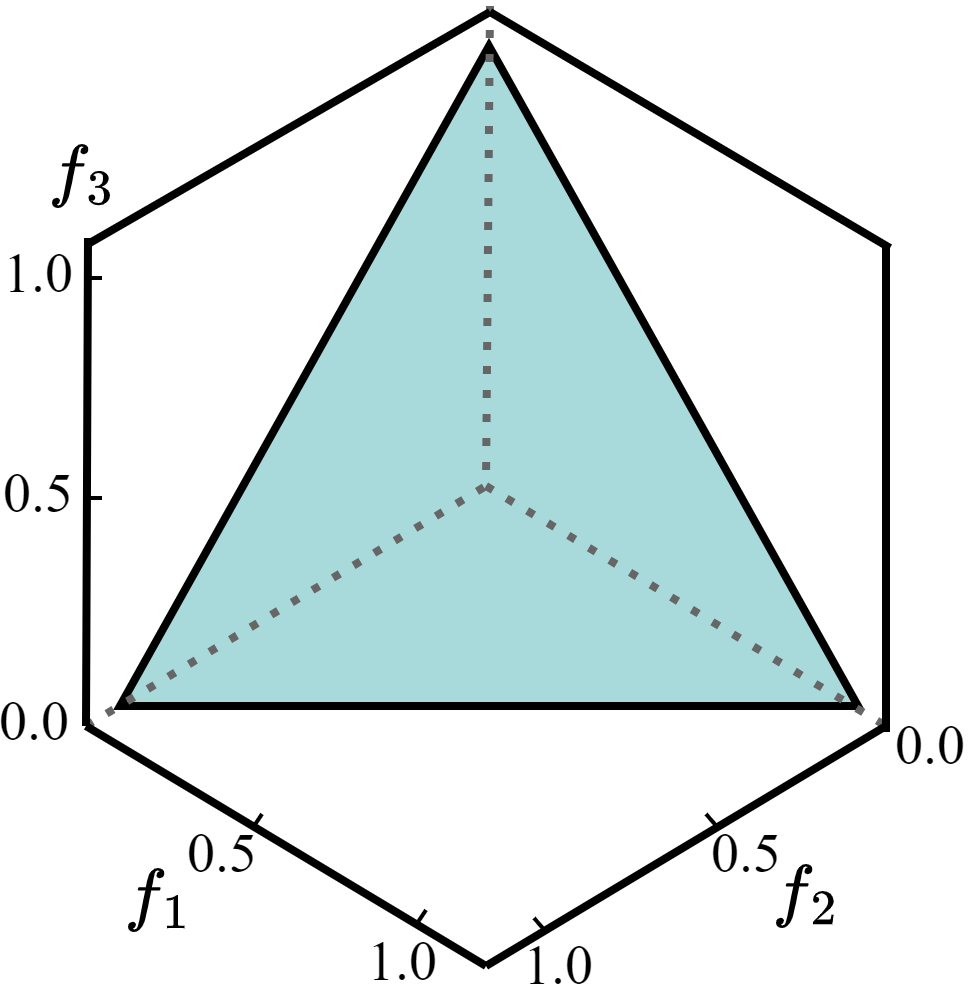}
        \caption{A special Pareto front in a 3-objective space }
        \label{mb}
    \end{subfigure}
        \hfill
    \begin{subfigure}[b]{0.21\textwidth}
        \centering
        \includegraphics[width=\linewidth]{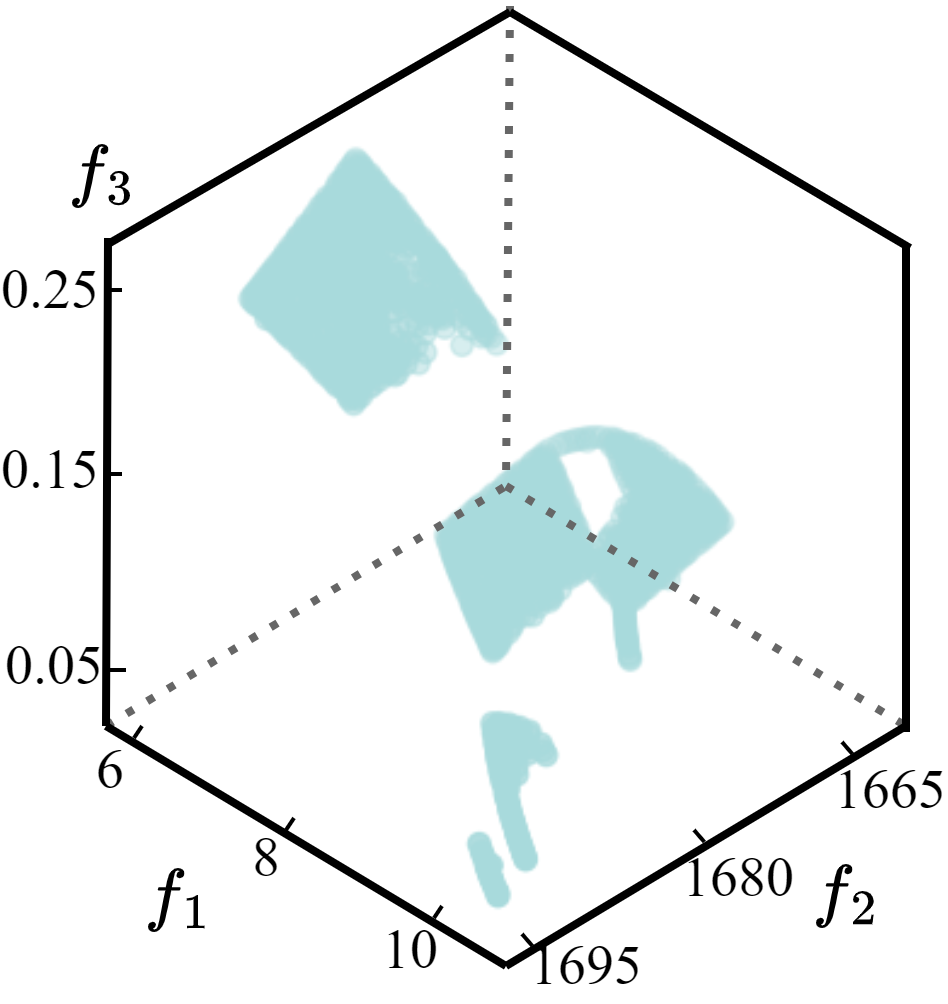}
        \caption{A realistic Pareto front in a 3-objective space}
        \label{md}
    \end{subfigure}
    \hfill
    \begin{subfigure}[b]{0.205\textwidth}
        \centering
        \includegraphics[width=\linewidth]{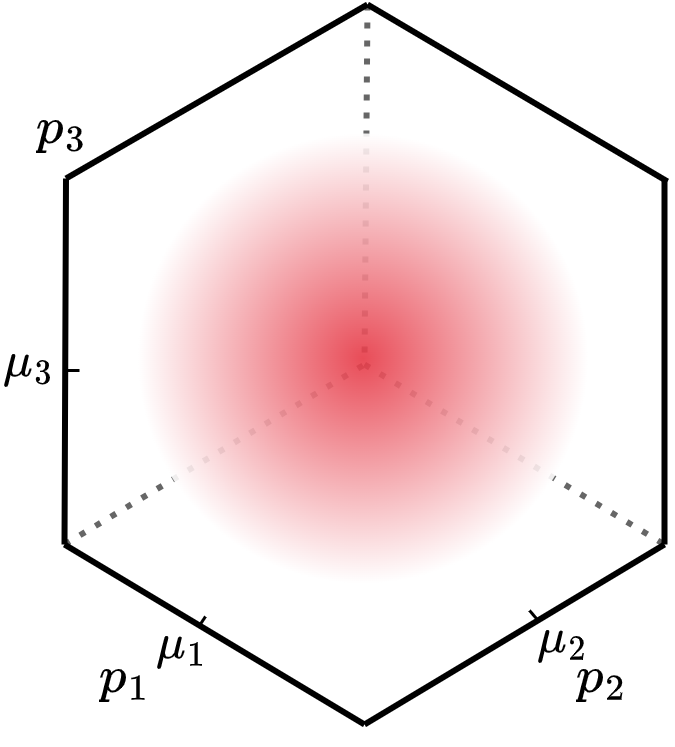}
        \caption{3D Gaussian sampling space}
        \label{mc}
    \end{subfigure}

    \caption{\textbf{Relation between the sampling distribution and the Pareto front shape.} 
    Fig. \ref{motiv}(a) shows that the preference sampling space is a triangular shape.
    Preference-based PSL methods work well when Pareto front shape (Fig. \ref{motiv}(b)) is triangular while do not match the Pareto front in Fig. \ref{motiv}(c). The proposed method does not require prior knowledge of the Pareto front shape and can approximate the Pareto set from any distribution, e.g., the 3D Gaussian distribution in Fig. \ref{motiv}(d).}
    \label{motiv}
    \vspace{-15pt} 
\end{figure*}

Recently, Pareto set learning (PSL) has been proposed to learn the Pareto set comprehensively \cite{lin2022pareto1,zhang2023hypervolume}. The basic idea of PSL is to convert a MOP into a single-objective problem through a preference vector \cite{lin2022pareto1} (we call preference-based PSL), and then optimize the converted single-objective problem. In particular, PSL employs a neural network to learn the mapping from preference vectors to Pareto optimal solutions, which holds notable advantages. Not only does it facilitate the learning of the complete Pareto set, but it can also generate a suitable Pareto optimal solution for an arbitrarily specified preference vector. Preference-based PSL algorithms have found applications in demanding real-world MOPs such as vehicle routing \cite{lin2022pareto1}, drug design \cite{jain2023multi} and model fairness \cite{ye2024praffl}.

From the geometrical perspective, preference-based PSL methods try to learn the intersecting point of each preference vector with the Pareto front by minimizing the weighted Techebycheff function (see Fig. \ref{fig:tch_goal}). In general, uniformly distributed preference vectors are used as the input to the neural network when there is no prior knowledge of the Pareto front shape. Unfortunately, the quality and efficiency of approximating the Pareto set are affected by the sampling distribution of the preference vectors \cite{chang2021self,ye2024data,ye2024evolutionary}. For example, Fig. 2(a) shows that the preference sampling space ($\sum_{i}p_{i}=1, 0 \leq p_i \leq 1, i = 1, 2, 3$), which works perfectly on the Pareto front in Fig. 2(b) but does not match well on the Pareto front in Fig. 2(c). Existing preference-based methods (using uniform preference vector sampling) are only effective for triangular Pareto fronts but not for other shapes. As a result, the Pareto set learned by a preference-based PSL method is not always a superior approximation of the true Pareto set. The sampling of preference vectors needs to adapt to the Pareto front shape of each MOP. However, the adaptation of preference sampling is difficult and time-consuming since it needs to estimate the Pareto front shape. 

To surmount the weaknesses of preference-based PSL and to make PSL effective in realistic MOPs, we propose a Pareto front shape-a\underline{g}nostic \underline{P}areto \underline{s}et \underline{l}earning (GPSL) that requires no preference vectors. Unlike preference-based PSL methods learning intersecting points between preference vectors and the Pareto front, we propose to maximize the \textit{similarity} (e.g., hypervolume indicator \cite{fonseca2006improved}) between the solution set generated by the neural network and the Pareto set. Additionally, GPSL can transform an arbitrary distribution into the Pareto set distribution. For example, GPSL samples from three-dimensional (3D) Gaussian space in Fig. 2(d) instead of triangular preference space (Fig. 2(a)). In this manner, the proposed method can learn the Pareto set without knowing the Pareto front shape.

The contributions of this paper can be summarized as follows:
\begin{itemize}

    \item To the best of our knowledge, the proposed GPSL is the first work to study Pareto front shape-agnostic Pareto set learning. This method focuses on maximizing the \textit{similarity} between the neural network-generated solution set and the true Pareto set. Notably, it enables sampling from any arbitrary distribution, instead of limiting it to a triangular preference distribution.
    \item 
    We propose two sampling distributions in GPSL, namely multivariate Gaussian sampling (GPSL-G) and Latin hypercube sampling (GPSL-L). GPSL-G and GPSL-L are not only applicable to triangular Pareto front but also to other complex Pareto fronts. 
    \item Extensive experiments demonstrate that the proposed GPSL can achieve better performance and faster convergence rate compared with state-of-the-art algorithms on three synthetic problems and nine real-world problems.
\end{itemize}

\section{Preliminary \label{psl}}
\subsection{Multi-Objective Optimization}
We consider the following multi-objective optimization problem (MOP) with $m$ objectives and $d$ decision variables
\begin{equation}
    \min_{\boldsymbol{x}\in\mathcal{X}}\boldsymbol{f}(\boldsymbol{x})=(f_1(\boldsymbol{x}),f_2(\boldsymbol{x}),\cdots,f_m(\boldsymbol{x})),
\end{equation}
where $\bm{x} \in \mathbb{R}^{d}$ is a d-dimensional decision vector (i.e., solution) in the decision space $\mathcal{X}$, $\bm{f}: \mathbb{R}^{d} \rightarrow \mathbb{R}^{m}$ is an objective function vector. We have the following definitions of multi-objective optimization:

\begin{definition}[Pareto Domination]
A solution $\bm{x}_{a}$ is said to dominate another solution $\bm{x}_{b}$, denoted as $\bm{x}_{a}\prec \bm{x}_{b}$, if and only if $\forall i \in \{1,\ldots,m\}$, $\bm{f}_{i}(\bm{x}_{a}) \leq  \bm{f}_{i}(\bm{x}_{b})$, and $\exists j \in \{1,\ldots,m\}, \bm{f}_{i}(\bm{x}_{a}) < \bm{f}_{i}(\bm{x}_{b})$.
\end{definition}
 
 \begin{definition}[Pareto optimal solution]
 Solution $\bm{x}_{a}$ is called a Pareto optimal solution if there does not exist a solution $\bm{x}_{b}$ such that $\bm{x}_{b} \prec \bm{x}_{a}$. In addition, $\bm{x}_{a}$ is said to weakly dominate $\bm{x}_{b}$, donated as $\bm{x}_{a} \preceq \bm{x}_{b}$ \textit{if and only if} $\forall i \in \{1,...,m\}$ such that $\bm{f}_{i}(\bm{x}_{a}) \leq \bm{f}_{i}(\bm{x}_{b})$.
 \end{definition}
 
\begin{definition}[Pareto Set and Pareto Front]
The set of all Pareto optimal solutions is called the \textit{Pareto set} (denoted as $\mathcal{M}_{ps}$). The \textit{Pareto front} (denoted as $\mathcal{P}(\mathcal{M}_{ps})$) is the image of the Pareto set in the objective space.
 \end{definition}


\subsection{Pareto Set Learning for Multi-Objective Optimization\label{PPSL}}

Recently, some studies have focused on utilizing the neural network to learn a continuous Pareto set. Pareto set learning samples preference vectors from a distribution. These preference vectors are input to the neural network, which outputs solutions for function evaluation. Finally, the aggregation function is used as the loss function to optimize the neural network \cite{navonlearning,lin2020controllable,shang2024collaborative}.

 A Pareto set model in Pareto set learning is defined as follows:
 \begin{equation}
     \bm{x}(\bm{p})=h_{\bm{\beta}}(\bm{p}),
 \end{equation}
where $\bm{p} \in \mathbb{R}^{m}$ is the preference vector ($\sum_{i}p_{i}=1, 0 \leq p_i \leq 1, i = 1, 2, 3)$) and $\bm{\beta}$ are parameters of the Pareto set model. The model maps any preference vector to its corresponding Pareto optimal solution. In the training phase of the Pareto set model, a preference vector $\bm{p}$ is generated and presented to the model. The output from the model is a solution $\bm{x}$. A scalarizing function is used to evaluate the quality of solution $\bm{x}$. The simplest way is to use weight-sum $g_{\mathrm{ws}}(\boldsymbol{x}|\boldsymbol{\bm{p}})=\sum_{i=1}^{m}p_{i}f_{i}(\bm{x})$. But weight-sum only works for the convex Pareto front \cite{boyd2004convex}. Usually, Pareto set learning uses the following weighted Tchebycheff function as a loss function
 \begin{equation} \label{agg}
    g_{\mathrm{tch}}(\boldsymbol{x}|\boldsymbol{\bm{p}})=\max_{1 \leq i \leq m} \{p_{i}(f_{i}(\bm{x})-(z_{i}^{*}-\epsilon))\}, 
 \end{equation}
 where $\bm{z}^{*}=({z}_{1}^{*}, z_{2}^{*},..., z_{m}^{*})$ is the ideal point in objective space (i.e., ${z_i}^{*}$ is the minimum value for each objective), and $\epsilon$ is a small positive value. The weighted Tchebycheff function in Equation(\ref{agg}) has a following property \cite{choo1983proper}:

 \begin{lemma} 
\label{tch}
A solution $\bm{x} \in \mathcal{X}$ is weakly Pareto optimal if and only if there is a preference vector $\bm{p}>0$ such that $\bm{x}$ is an optimal solution of the minimization problem of the weighted Tchebycheff function.
\end{lemma}



According to Lemma \ref{tch}, any corresponding Pareto optimal solution in the Pareto set can be learned by optimizing Equation (\ref{agg}) for a corresponding preference vector. The optimization task of the Pareto set model can be expressed as follows:
\begin{equation} \label{psl}
    \boldsymbol{\beta}^*=\underset{\boldsymbol{\beta}}{\operatorname*{\arg\min}} \ \mathbb{E}_{\bm{p} \sim \mathcal{D}(\cdot)}g_{\mathrm{tch}}(h_{\boldsymbol{\beta}}(\boldsymbol{p})|\boldsymbol{p}),
\end{equation}
 where $\bm{p}$ is sampled from the Dirichlet distribution $\mathcal{D}(\cdot)$. The Monte Carlo sampling is used to approximate the expectation term in Equation (\ref{psl}) as follows: 
 \begin{equation}
         \boldsymbol{\beta}^*=\underset{\boldsymbol{\beta}}{\operatorname*{\arg\min}} \ \frac{1}{N}\sum_{i=1}^N  g_{\mathrm{tch}}(h_{\boldsymbol{\beta}}(\boldsymbol{p}^{(i)})|\boldsymbol{p}^{(i)}),
 \end{equation}
 where  $\bm{p}^{(i)} \sim \mathcal{D}(\cdot)$. The gradient descent algorithm can be used to optimize the Pareto set model as follows:
 \begin{equation}
     \boldsymbol{\beta}_{t+1}=\boldsymbol{\beta}_t-\eta \frac{1}{N}\sum_{i=1}^N\nabla_{\boldsymbol{\beta}}g_{\mathrm{tch}}(h_{\boldsymbol{\beta}}(\boldsymbol{\bm{p}^{(i)}})|\boldsymbol{\bm{p}^{(i)}}),
 \end{equation}
where $\eta$ is the learning rate. After enough training, the Pareto set model can output a (near) Pareto optimal solution corresponding to an arbitrary preference vector. 


\section{Pareto Front Shape-Agnostic Pareto Set Learning}

Preference-based PSL methods aim to map a preference vector into the related optimal solution by optimizing the aggregation function. The performance of the learned Pareto set is affected by the quality of sampling preference vectors \cite{chen2022multi,chang2021self}. To address this issue, we consider a method that does not require a specific preference vector space (e.g., triangle in Fig. \ref{motiv}(a)). Such a method can generate a Pareto set distribution from any distribution instead of using the preference vector space. The remainder of this section will introduce PSL as a distribution transformation problem and consider how to optimize it.

\subsection{Pareto Set Learning via Distribution Transformation}
We consider Pareto set learning using distribution transformation, which is defined as follows:

\begin{figure}[t]
    \centering
    \includegraphics[width=\linewidth]{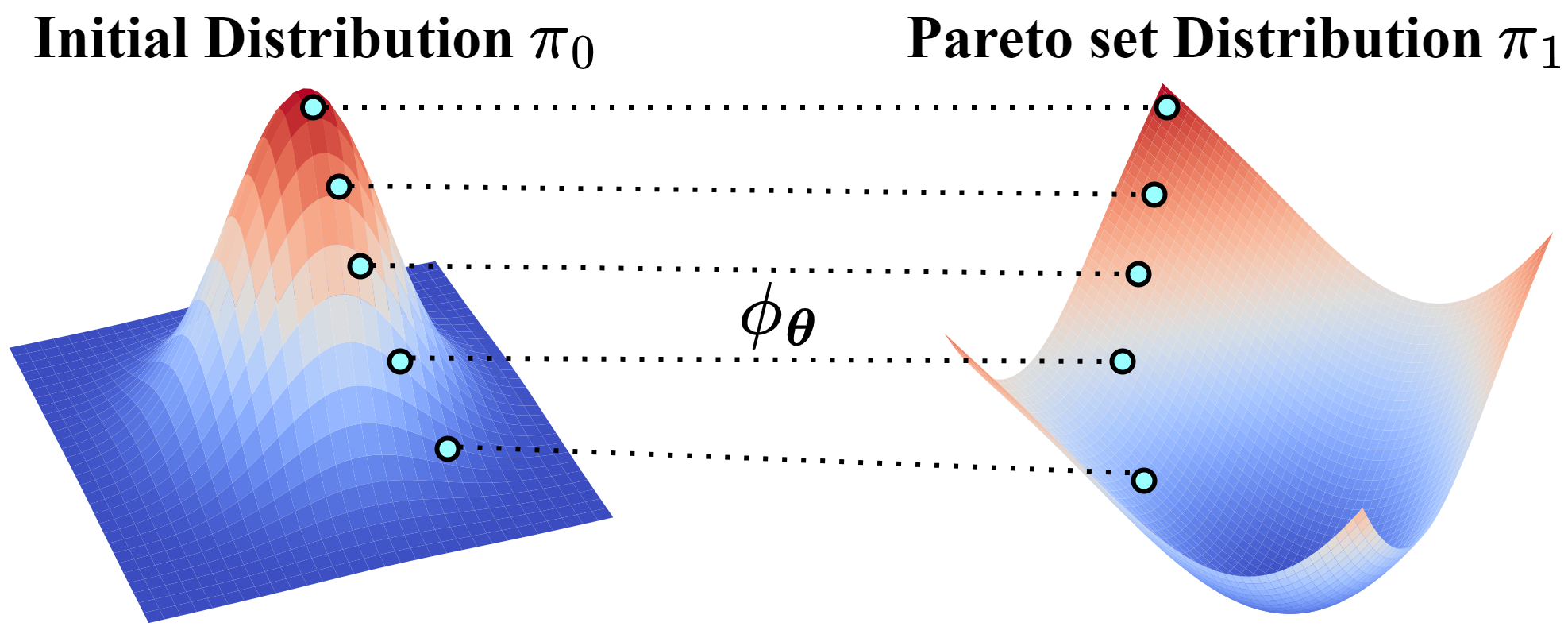}
    \caption{\textbf{The learning goal of Pareto front shape-agnostic Pareto set learning (GPSL).} GPSL transforms an initial distribution (e.g., Gaussian distribution) to the Pareto set distribution. During training, the model $\phi_{\bm{\theta}}$ tries to find the mapping from each sampling point in the initial distribution to the Pareto optimal solution.
}   
    \label{fig:diff}
    \vspace{-0.4cm}
\end{figure}

\begin{definition}[Distribution Transformation]
\label{def:trans} 
Let the distribution of the true Pareto set be denoted as $\pi_{1} \in \mathbb{R}^{d}$ and an initial distribution be denoted as $\pi_{0}  \in \mathbb{R}^{k}$ ($k \geq 1$). The goal of the model $\phi_{\bm{\theta}}$: $\mathbb{R}^{k} \rightarrow \mathbb{R}^d$ is to transform $\pi_{0}$ to $\pi_{1}$. The model $\phi_{\bm{\theta}}$ is an optimal transformation if $\phi_{\bm{\theta}}(\bm{v})=u$, for $\forall \bm{u} \in \pi_{1}, {\exists} \bm{v} \in \pi_{0}$.
\end{definition}

 As shown in Fig. \ref{fig:diff}, the model $\phi_{\bm{\theta}}$ can eventually obtain the distribution of the true Pareto set from $\pi_{0}$. Although the true Pareto set is typically unknown, some indicators in multi-objective optimization can measure the \textit{similarity} between the solution set learned by the model and the true Pareto set. For example, we can use the hypervolume indicator \cite{fonseca2006improved}, which is defined as follows:

\begin{definition}[Hypervolume]
The hypervolume indicator $\mathcal{H}_{\bm{r}}(\mathcal{Y})$ of a set of points $\mathcal{Y} \subset \mathbb{R}^{m}$ is the $m$-dimensional Lebesgue measure of the region dominated by $\mathcal{P}(\mathcal{Y})$ (Pareto front) and bounded by the reference point $\bm{r} \in \mathbb{R}^{m}$.
\end{definition}

We directly use hypervolume as the optimization goal of the model $\phi_{\bm{\theta}}$, then we can get the Pareto set when hypervolume is maximized, as shown by following proposition \cite{fleischer2003measure}:

\begin{proposition}[(hypervolume maximum $\Longrightarrow$ Pareto Set] \label{hvmm}
Let $\bm{r}$ be a reference point that is dominated by any point on dominates the whole Pareto front $\mathcal{P}$, and $B$ be a non-dominated solution set. Then $B^{*}$ is the Pareto set if $B^{*}=\text{arg max}_{B} \mathcal{H}_{\bm{r}}(\bm{f}(B))$.
\end{proposition}

Based on Proposition \ref{hvmm}, $\phi_{\bm{\theta}}$ can convert distribution $\pi_{0}$ to the Pareto set distribution $\pi_{1}$ by maximizing hypervolume. $\pi_{0}$ can be initialized to any distribution (such as Gaussian distribution, uniform distribution, etc.). The learning goal of the model $\phi_{\bm{\theta}}$ is as follows:
\begin{equation}
    \bm{\theta}^{*} = {\arg \max}_{\bm{\theta}} \mathcal{H}_r (\phi_{\bm{\theta}} (\bm{v})),
\end{equation}
where $\bm{v}$ is sampled from $\pi_{0}$ ($\bm{v} \sim \pi_{0}$). Then, we can obtain the Pareto set distribution by $\widetilde{\pi}_{1}=\phi_{\bm{\theta}^{*}}(\pi_{0})$. Our novel contribution is to handle the Pareto set learning task as a distribution transformation problem with hypervolume maximization.

\subsection{Optimizing GPSL}
One difficulty of the hypervolume indicator is the exponential increase of its calculation time with the number of objectives. To avoid spending impractically long computation time by trying to calculate the exact hypervolume value of a solution set for a multi-objective problem with many objectives,  we use the following R2-based approximation method for approximate hypervolume calculation:

\begin{figure}[t]
  \centering
\includegraphics[width=\linewidth]{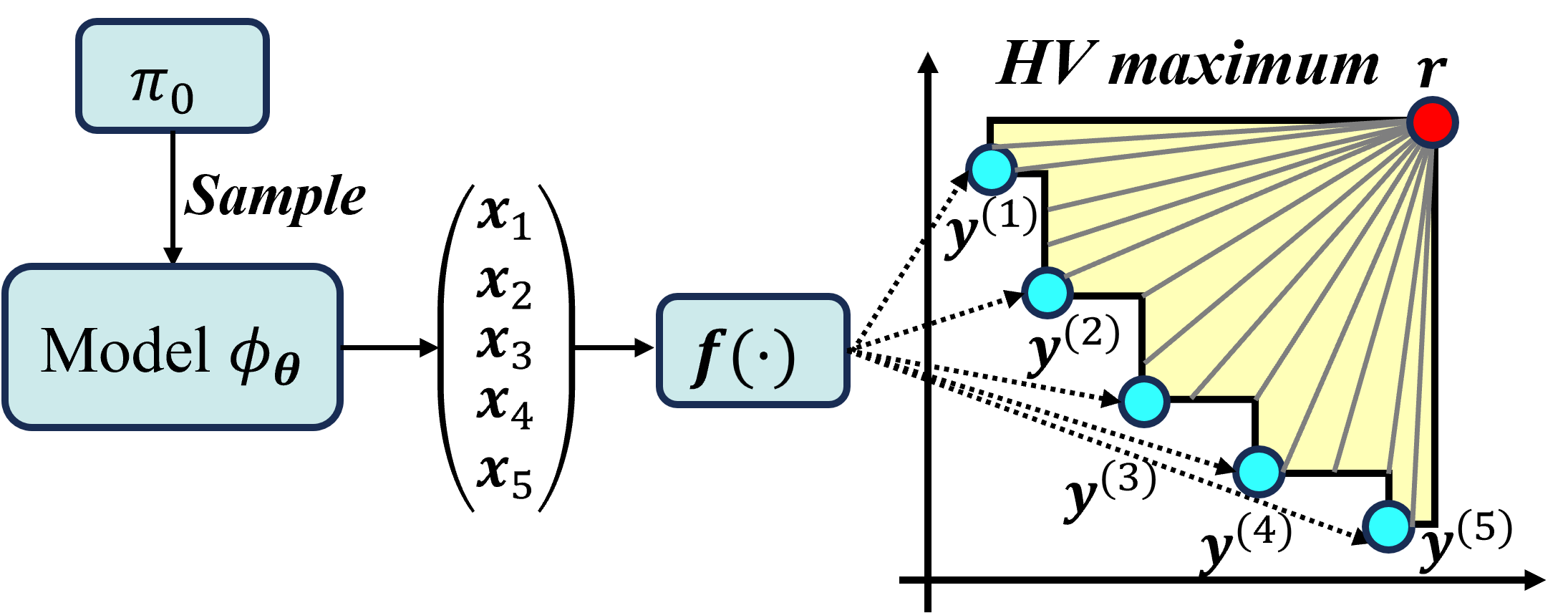}
  \caption{\textbf{The process of model training.} The yellow area is the exact hypervolume value. The sum of the lengths of the line segments is used to approximate the exact hypervolume of the yellow area.}
  \label{hvm}
  \vspace{-0.4cm}
\end{figure}

\begin{definition}[R2-based hypervolume approximation \cite{shang2018new}]
\label{maxhv}
Let $B=\{\bm{y}^{(1)}, \bm{y}^{(2)},...,\bm{y}^{(N)}\}$ be a set of solutions in the objective space and $\bm{r}$ is a reference point, which satisfies $\bm{y}^{(i)} \prec \bm{r} (i=1,2,..., N)$. The approximated hypervolume of $B$ can be expressed as follows:
\begin{equation} 
\widetilde{\mathcal{H}_{r}}(B)= c_{m} \sum_{\bm{\lambda} \in\Lambda} \left\{\max_{\bm{y}\in B}  \min_{1 \leq i \leq m} \frac{r_{i}-y_{i}}{\lambda_{i}}\right\}^m, 
\label{lem}
\end{equation}
where \(\Lambda\) represents the set of direction vectors, $\bm{\lambda}=(\lambda_{1}, \lambda_{2},...,\lambda_{m}) \in \Lambda$ is a direction vector with $||\bm{\lambda}||_2=1$ and $\lambda_{i} \geq 0$. \( c_m \)$=\frac{\pi^{m/2}}{m|\Lambda|2^{m-1}\Gamma(m/2)}$ denotes a constant which only depends on $m$, and $\Gamma(\cdot)$ is the Gamma function. 
\end{definition}

Among Equation \ref{lem}, $\min_{1 \leq i \leq m} \left\{\frac{r_{i}-y_{i}}{\lambda_{i}}\right\}$ is a modified Techebycheff aggregation function. The geometric meaning of the modified Techebycheff aggregation function is the minimal projected distance from the vector $\bm{y}^{(i)} \in B$ to the vector $\bm{\lambda} \in \Lambda$ with the reference point $\bm{r}$ \cite{zhang2023hypervolume}. Computing the approximated hypervolume requires a set of direction vectors $\Lambda$. Das and Dennis’s method \cite{das1998normal} is the most commonly used systematic method for uniformly distributed reference point sampling. In this work, we also use Das and Dennis’s method. 

 \begin{figure*}[t]
    \centering
    \includegraphics[width=0.95\linewidth]{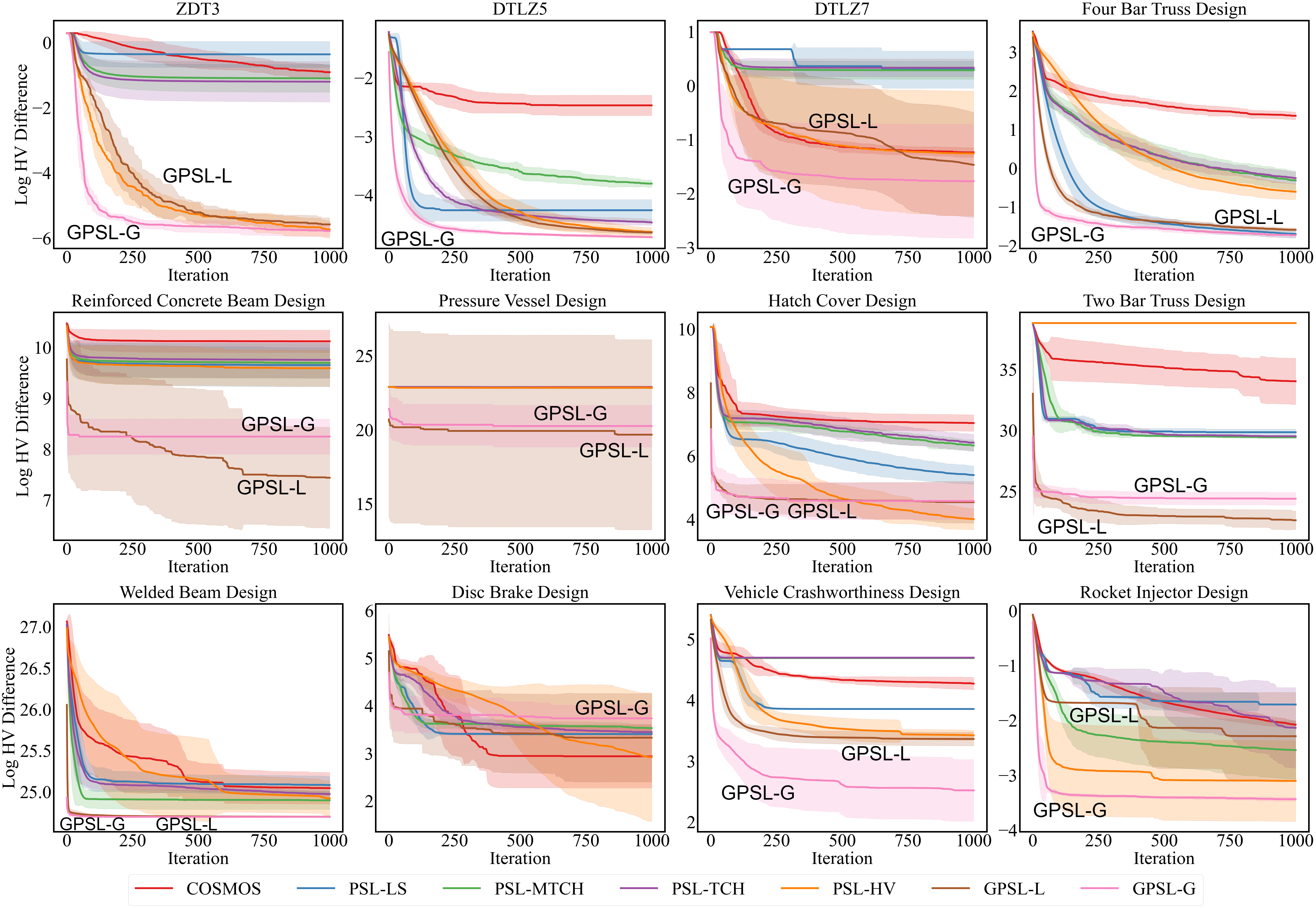}
    \caption{\textbf{The log HV difference on 12 different problems.} The solid line and shaded area are the mean and standard deviation of 11 independent runs of each algorithm, respectively.}
    \label{fig:result}
    \vspace{-0.4cm}
\end{figure*}



Hypervolume maximization drives the transformation of any distribution into the Pareto set distribution. Fig. \ref{hvm} shows the training of GPSL. First, a batch of data points are sampled from $\pi_{0}$ and input into the model. Then, the model transforms these data points to the Pareto optimal solutions by maximizing hypervolume. The optimization goal can be expressed in terms of an expectation
\begin{equation}
    \min_{\bm{\theta} \in \bm{\Theta}} \mathcal{L}(\bm{\theta}) = - \ \mathbb{E}_{\bm{v} \sim \pi_{0}}[ \widetilde{\mathcal{H}_{\bm{r}}}(\bm{f}(\phi_{\bm{\theta}}(\bm{v})))].
\end{equation}
It is difficult to solve this optimization problem due to the expectation term. We use Monte Carlo sampling to estimate $\mathcal{L}(\bm{\theta})$, as follows:
\begin{equation}
    \widetilde{\mathcal{L}}(\bm{\theta}) = -\frac{1}{N} \sum_{i=1}^{N} \widetilde{\mathcal{H}_{\bm{r}}}(\bm{f}(\phi_{\bm{\theta}}(\bm{v}^{(i)})|\bm{v}^{(i)}),
    \label{eq10}
\end{equation}
where $N$ is the number of sampling data and $\bm{v}^{(i)}$ is sampled from distribution $\pi_{0}$ independently. A crucial step is to find a gradient direction to update the model at each iteration. The gradient of the $\widetilde{\mathcal{H}_{\bm{r}}}(\bm{\theta})$ can be calculated by chain rule

 \begin{equation}
     \nabla_{\bm{\theta}}\widetilde{\mathcal{H}_{\bm{r}}}(\bm{\theta})=\frac{\partial \phi_{\bm{\theta}}(\bm{v})}{\partial \theta} \cdot \nabla_{\bm{x}}\widetilde{\mathcal{H}_{\bm{r}}}(\bm{f}(\bm{x}))_{|\bm{x}=\phi_{\bm{\theta}}(\bm{v})},
 \end{equation}
where $\nabla_{\bm{x}}\widetilde{\mathcal{H}_{\bm{r}}}(\bm{f}(\bm{x}))_{|\bm{x}=\phi_{\bm{\theta}}(\bm{v})}$ is the gradient of approximated hypervolume for decision variables $\bm{x}$. $\frac{\partial \phi_{\bm{\theta}}(B)}{\partial \bm{\theta}}$ can be easily calculated with backpropagation in the neural network. Then, the gradient of $\widetilde{\mathcal{L}}(\bm{\theta})$ can be calculated as:
\begin{equation}
    \bar{\nabla}_{\bm{\theta}}\widetilde{\mathcal{L}}(\bm{\theta})= -\frac{1}{N} \sum_{i=1}^{N} \frac{\partial \phi_{\bm{\theta}}(\bm{v}^{(i)})}{\partial \bm{\theta}} \cdot \nabla_{\bm{x}}\widetilde{\mathcal{H}_{r}}(\bm{f}(\bm{x}))_{|\bm{x}=\phi_{\bm{\theta}}(\bm{v}^{(i)})}.
\end{equation}

Finally, we can use gradient descent to optimize the parameters of the model $\phi_{\bm{\theta}}$ as follows:
\begin{equation} \label{update}
    \bm{\theta}_{t+1} = \bm{\theta}_{t} - \eta \bar{\nabla}_{\bm{\theta}}\widetilde{\mathcal{L}}(\bm{\theta}),
\end{equation}
 where $\eta$ is the learning rate of the neural network and the $t$ represents $t$-th iteration. In our work, $\pi_{0}$ (i.e., the initial distribution) can be any distribution. We tried Gaussian distribution sampling and Latin hypercube sampling, called GPSL-G and GPSL-L, respectively. 
 

  It is worth noting that the idea behind our proposed GPSL is from a perspective: why using preference vectors to learn the Pareto set is inefficient (because most of the Pareto front shapes of multi-objective optimization problems are not triangular). Furthermore, we generalize the input to any distribution to improve the learning efficiency of the Pareto set.

  
  

\section{Experiments}

This section first compares the performance of GPSL with other advanced methods on synthetic and real-world problems. 
Moreover, we analyze the impact of different distributions and dimensions of GPSL on the approximated Pareto front.

\subsection{Experimental Settings}
\subsubsection{Baseline Algorithms}

We compare five advanced algorithms including PSL-LS \cite{navonlearning}, PSL-TCH \cite{tuan2024framework}, PSL-MTCH \cite{lin2022pareto2}, COSMOS \cite{ruchte2021scalable}, and PSL-HV \cite{zhang2023hypervolume}. The difference between these five algorithms lies in different aggregation functions. All algorithms are implemented in PyTorch \cite{paszke2019pytorch}, and the Adam optimizer \cite{kingma2014adam} is used to train all algorithms. We evaluate all algorithms using the log hypervolume (HV) difference between the true and learned Pareto front. The calculation method of log HV difference is expressed as follows:
\begin{equation}
    \text{log HV difference} = log (HV + \epsilon - \Hat{HV}),
\end{equation}
where HV is the hypervolume of the true/approximated Pareto front, $\Hat{HV}$ is the hypervolume of the learned Pareto front by each algorithm and $\epsilon \geq 0$ is a small value.

\subsubsection{Benchmarks and Real-World Problems}
The algorithms are first compared on three synthetic test problems (ZDT3 \cite{zitzler2000comparison}, DTLZ5 \cite{deb2005scalable}, and DTLZ7 \cite{deb2005scalable}) implemented in pymoo\footnote{https://pymoo.org/problems/index.html}. Then, we also conduct experiments on nine real-world multi-objective engineering design problems\footnote{https://github.com/ryojitanabe/reproblems} \cite{tanabe2020easy}. Since the true Pareto fronts of these real-world problems are unknown, we use the approximated Pareto fronts provided by the problem proposers. All experiments are tested on a server with Intel(R) Xeon(R) Gold 6138 CPU@2.00 GHz and a GeForce RTX 2080 Ti GPU (11GB RAM).

\subsubsection{Parameter Settings} 
The dimension of the initial distribution is set to the number of decision variables $d$. The sampling center of the proposed GPSL-G and GPSL-L algorithms is $(lb+ub)/2$, where $lb$ and $up$ are the upper and lower bounds of the decision variables in each test problem, respectively. The maximum iterations are 1000, and the batch size (in Equation (\ref{eq10})) is 32. All algorithms are run 11 times on each test problem. The same number of data points are input to each model to fairly evaluate all algorithms. Thus, each model can produce the same number of solutions.

\begin{figure*}[t] 
  \centering
\includegraphics[width=0.95\linewidth]{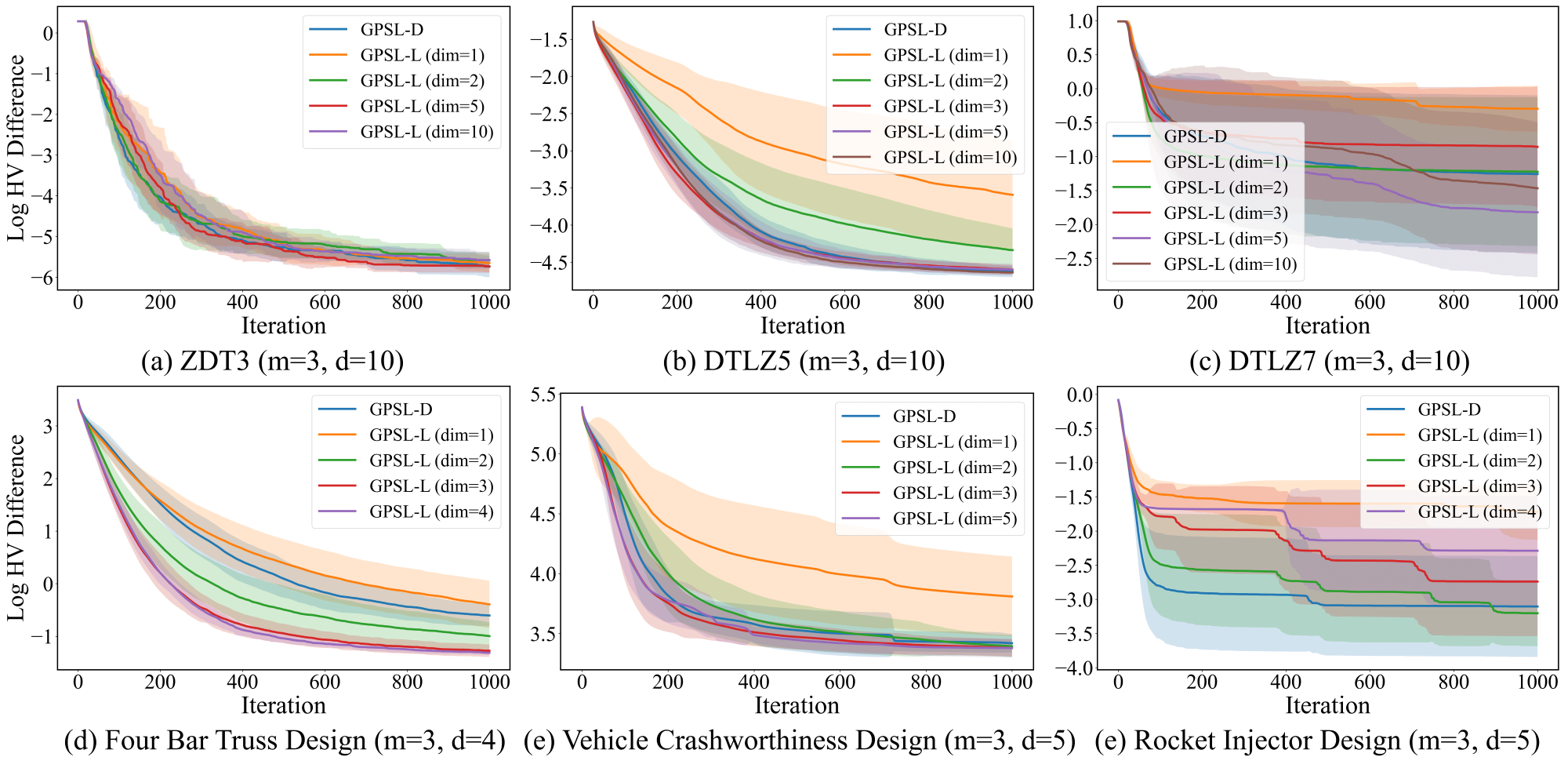}
  \caption{The effect of different dimensions for GPSL-G on six test problems over 11 runs.}
  \label{dim_g}
  \vspace{-0.4cm}
\end{figure*}

\begin{figure*}[!htb] 
  \centering
\includegraphics[width=0.95\linewidth]{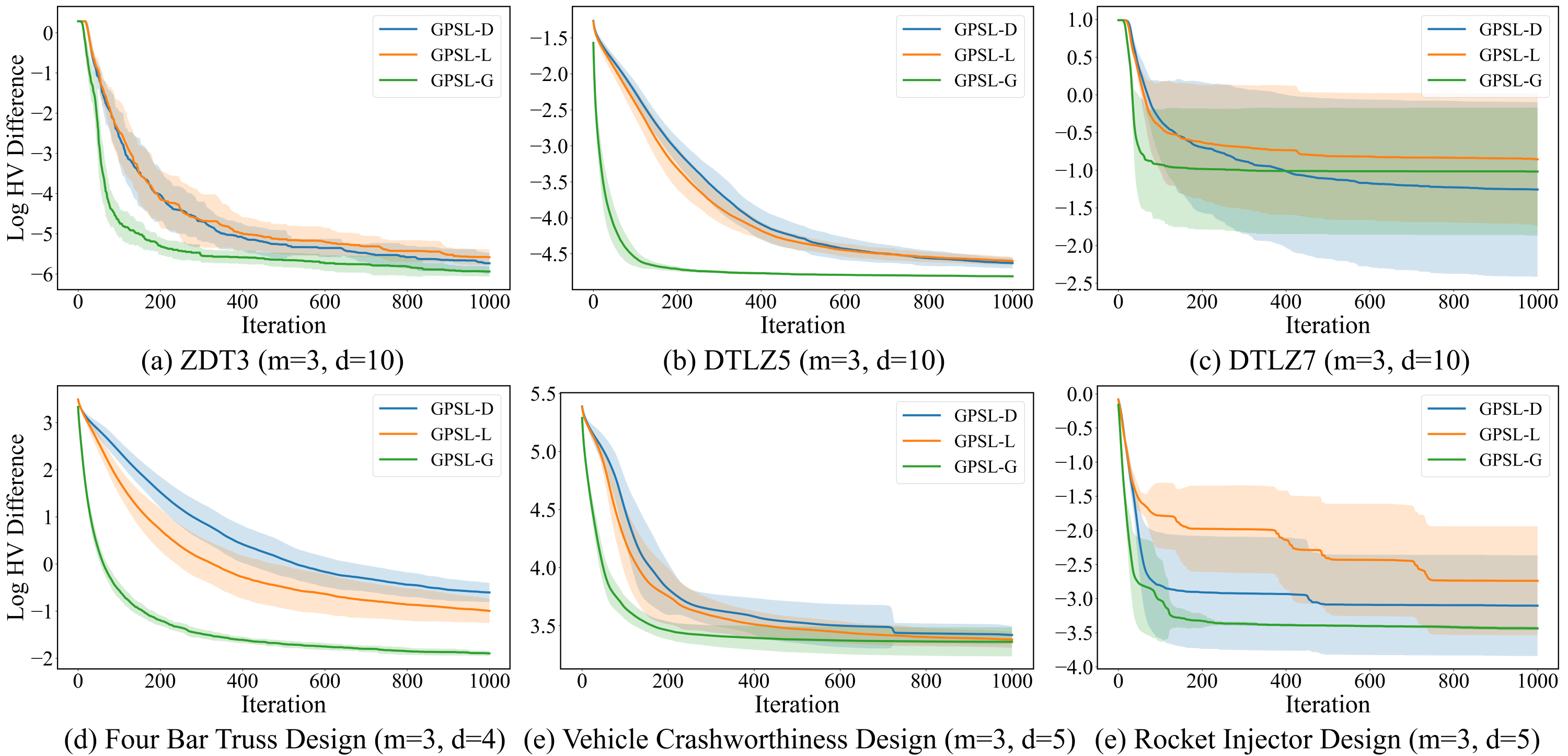}
  \caption{The effect of different distributions on six test problems over 11 runs (the dimension for each distribution is equal to $m$).}
  \label{dist}
  \vspace{-0.4cm}
\end{figure*}


\subsection{Main Results}

We compare the solution sets obtained by GPSL-L, GPSL-G, and the other preference-based PSL methods on each of the 12 test problems. Fig. \ref{fig:result} shows the log HV difference to the true/approximated Pareto front for each of the 12 test problems during training. In most test problems, our proposed Pareto front shape-agnostic PSL algorithms (GPSL-L and GPSL-G) perform better than the other preference-based PSL algorithms. Especially, our proposed GPSL-G shows faster convergence than the other algorithms on almost all test problems. One reason for the better performance of our proposed GPSL is that all test problems have irregular Pareto fronts. As pointed out in \cite{chen2022multi}, Pareto set learning is challenging when the Pareto front shape is irregular. Our experimental results clearly show the high ability of the proposed algorithms to handle irregular Pareto fronts. One exception in Fig. \ref{fig:result} is the Disc brake design problem (i.e., the second figure from the left in the bottom row). 
The third objective of this problem is the total constraint violation. That is, this three-objective problem has been created from a constrained two-objective problem by using the total constraint violation as the third objective. The scale of the total constraint violation is different from the other two objectives. This problem characteristic poses a challenge to all algorithms. The referenced Pareto front has a large value in the third objective, and it is hard to set a suitable reference point for the hypervolume maximum.


\subsection{Ablation Studies}
In this section, we analyze the effects of the dimensionality (dim) of the initial distribution and the type of the initial distribution on the performance of our proposed GPSL. Fig. \ref{dim_g} shows the performance of GPSL-G ($\mathcal{N}(\bm{0}, I)$) in different dimensions. GPSL-D represents sampling in the Dirichlet distribution, which is the same as the preference vector sampling distribution, and its dimension is equal to the objective space dimension $m$. We can observe that smaller dim settings (e.g., dim=2) tend to converge faster. This is because the neural network learns faster in the small-dimensional sampling space than in the high-dimensional sampling space. This conclusion can be clearly reflected in the simulation of DTLZ5 and the Four Bar Truss Design problem where curves with larger dim converge slower. However, the representation ability of the sampling space with a low dim (e.g., dim=1) is insufficient resulting in poor Pareto front approximation (i.e., poor convergence results). This conclusion can be clearly reflected in the simulation of ZDT3 and DTLZ7 problems where the convergence results of the curve with dim=1 are worse than other curves. Thus, the effectiveness of GPSL on complex multi-objective optimization problems remains stable when the dimensionality of the sampling space is greater than or equal to 2.

Fig. \ref{dist} compares the three distributions: Dirichlet distribution (GPSL-D), Latin hypercube (GPSL-L), and Gaussian distribution (GPSL-G). For almost all test problems, the best results are obtained from Gaussian distribution. This may be because the Gaussian distribution has no constraints in comparison with the Dirichlet distribution on the hyperplane and Latin hypercube with a special rule. 


\section{CONCLUSIONS}
In this paper, we formulated Pareto set learning as a transformation problem from an arbitrary distribution into the Pareto set distribution. We used hypervolume maximization as the optimization goal to optimize the model. Theoretically, a solution set is the Pareto optimal set when the hypervolume is maximized. Our work provides a new perspective to learn the Pareo set, that is, the input of the neural network can be an arbitrary distribution rather than the fixed preference vector distribution. We implemented two algorithms using Gaussian distribution (GPSL-G) and Latin hypercube distribution (GPSL-L). Our experimental results on artificial test problems and real-world problems clearly showed that GPSL-L and GPSL-G outperformed the five state-of-the-art preference-based algorithms on most problems. Our experimental results suggested that the traditional PSL methods are insufficient on the non-triangular Pareto front shape, while the convergence speed and convergence results of our proposed method have obvious advantages.

\bibliographystyle{IEEEtran}
\bibliography{IEEEabrv,mybibfile}

\begin{thebibliography}{10}
\providecommand{\url}[1]{#1}
\csname url@samestyle\endcsname
\providecommand{\newblock}{\relax}
\providecommand{\bibinfo}[2]{#2}
\providecommand{\BIBentrySTDinterwordspacing}{\spaceskip=0pt\relax}
\providecommand{\BIBentryALTinterwordstretchfactor}{4}
\providecommand{\BIBentryALTinterwordspacing}{\spaceskip=\fontdimen2\font plus
\BIBentryALTinterwordstretchfactor\fontdimen3\font minus \fontdimen4\font\relax}
\providecommand{\BIBforeignlanguage}[2]{{%
\expandafter\ifx\csname l@#1\endcsname\relax
\typeout{** WARNING: IEEEtran.bst: No hyphenation pattern has been}%
\typeout{** loaded for the language `#1'. Using the pattern for}%
\typeout{** the default language instead.}%
\else
\language=\csname l@#1\endcsname
\fi
#2}}
\providecommand{\BIBdecl}{\relax}
\BIBdecl

\bibitem{tanabe2020easy}
R.~Tanabe and H.~Ishibuchi, ``An easy-to-use real-world multi-objective optimization problem suite,'' \emph{Applied Soft Computing}, vol.~89, p. 106078, 2020.

\bibitem{liao2008multiobjective}
X.~Liao, Q.~Li, X.~Yang, W.~Zhang, and W.~Li, ``Multiobjective optimization for crash safety design of vehicles using stepwise regression model,'' \emph{Structural and Multidisciplinary Optimization}, vol.~35, pp. 561--569, 2008.

\bibitem{ashby2000multi}
M.~Ashby, ``Multi-objective optimization in material design and selection,'' \emph{Acta Materialia}, vol.~48, no.~1, pp. 359--369, 2000.

\bibitem{ehrgott2005multicriteria}
M.~Ehrgott, \emph{Multicriteria Optimization}.\hskip 1em plus 0.5em minus 0.4em\relax Springer Science \& Business Media, 2005, vol. 491.

\bibitem{miettinen1999nonlinear}
K.~Miettinen, \emph{Nonlinear Multiobjective Optimization}.\hskip 1em plus 0.5em minus 0.4em\relax Springer Science \& Business Media, 1999, vol.~12.

\bibitem{zitzler1999evolutionary}
E.~Zitzler, \emph{Evolutionary Algorithms for Multiobjective Optimization: Methods and Applications}.\hskip 1em plus 0.5em minus 0.4em\relax Shaker Ithaca, 1999, vol.~63.

\bibitem{zhou2011multiobjective}
A.~Zhou, B.-Y. Qu, H.~Li, S.-Z. Zhao, P.~N. Suganthan, and Q.~Zhang, ``Multiobjective evolutionary algorithms: A survey of the state of the art,'' \emph{Swarm and Evolutionary Computation}, vol.~1, no.~1, pp. 32--49, 2011.

\bibitem{hillermeier2001generalized}
C.~Hillermeier, ``Generalized homotopy approach to multiobjective optimization,'' \emph{Journal of Optimization Theory and Applications}, vol. 110, no.~3, pp. 557--583, 2001.

\bibitem{zhang2008rm}
Q.~Zhang, A.~Zhou, and Y.~Jin, ``{RM-MEDA}: A regularity model-based multiobjective estimation of distribution algorithm,'' \emph{IEEE Transactions on Evolutionary Computation}, vol.~12, no.~1, pp. 41--63, 2008.

\bibitem{lin2022pareto1}
X.~Lin, Z.~Yang, and Q.~Zhang, ``{Pareto} set learning for neural multi-objective combinatorial optimization,'' \emph{arXiv preprint arXiv:2203.15386}, 2022.

\bibitem{zhang2023hypervolume}
X.~Zhang, X.~Lin, B.~Xue, Y.~Chen, and Q.~Zhang, ``Hypervolume maximization: A geometric view of {Pareto} set learning,'' \emph{In Proceedings of Advances in Neural Information Processing Systems}, vol.~36, pp. 38\,902--38\,929, 2023.

\bibitem{jain2023multi}
M.~Jain, S.~C. Raparthy, A.~Hern{\'a}ndez-Garc{\i}a, J.~Rector-Brooks, Y.~Bengio, S.~Miret, and E.~Bengio, ``Multi-objective gflownets,'' \emph{In Proceedings of International Conference on Machine Learning}, pp. 14\,631--14\,653, 2023.

\bibitem{ye2024praffl}
R.~Ye and M.~Tang, ``{PraFFL}: A preference-aware scheme in fair federated learning,'' \emph{arXiv preprint arXiv:2404.08973}, 2024.

\bibitem{chang2021self}
S.~Chang, K.~Yoo, J.~Jang, and N.~Kwak, ``Self-evolutionary optimization for {Pareto} front learning,'' \emph{arXiv preprint arXiv:2110.03461}, 2021.

\bibitem{ye2024data}
R.~Ye, L.~Chen, W.~Liao, J.~Zhang, and H.~Ishibuchi, ``Data-driven preference sampling for {Pareto} front learning,'' \emph{arXiv preprint arXiv:2404.08397}, 2024.

\bibitem{ye2024evolutionary}
R.~Ye, L.~Chen, J.~Zhang, and H.~Ishibuchi, ``Evolutionary preference sampling for {Pareto} set learning,'' \emph{In Proceedings of the Genetic and Evolutionary Computation Conference}, 2024.

\bibitem{fonseca2006improved}
C.~M. Fonseca, L.~Paquete, and M.~L{\'o}pez-Ib{\'a}nez, ``An improved dimension-sweep algorithm for the hypervolume indicator,'' \emph{In Proceedings of International Conference on Evolutionary Computation}, pp. 1157--1163, 2006.

\bibitem{navonlearning}
A.~Navon, A.~Shamsian, E.~Fetaya, and G.~Chechik, ``Learning the {Pareto} front with hypernetworks,'' \emph{In Proceedings of International Conference on Learning Representations}, 2021.

\bibitem{lin2020controllable}
X.~Lin, Z.~Yang, Q.~Zhang, and S.~Kwong, ``Controllable {Pareto} multi-task learning,'' \emph{arXiv preprint arXiv:2010.06313}, 2020.

\bibitem{shang2024collaborative}
C.~Shang, R.~Ye, J.~Jiang, and F.~Gu, ``Collaborative {Pareto} set learning in multiple multi-objective optimization problems,'' \emph{arXiv preprint arXiv:2404.01224}, 2024.

\bibitem{boyd2004convex}
S.~Boyd and L.~Vandenberghe, \emph{Convex optimization}.\hskip 1em plus 0.5em minus 0.4em\relax Cambridge University Press, 2004.

\bibitem{choo1983proper}
E.~U. Choo and D.~R. Atkins, ``Proper efficiency in nonconvex multicriteria programming,'' \emph{Mathematics of Operations Research}, vol.~8, no.~3, pp. 467--470, 1983.

\bibitem{chen2022multi}
W.~Chen and J.~Kwok, ``Multi-objective deep learning with adaptive reference vectors,'' \emph{In Proceedings of Advances in Neural Information Processing Systems}, vol.~35, pp. 32\,723--32\,735, 2022.

\bibitem{fleischer2003measure}
M.~Fleischer, ``The measure of {Pareto} optima applications to multi-objective metaheuristics,'' \emph{In Proceedings of International Conference on Evolutionary Multi-criterion Optimization}, pp. 519--533, 2003.

\bibitem{shang2018new}
K.~Shang, H.~Ishibuchi, M.-L. Zhang, and Y.~Liu, ``A new r2 indicator for better hypervolume approximation,'' \emph{In Proceedings of Proceedings of the Genetic and Evolutionary Computation Conference}, pp. 745--752, 2018.

\bibitem{das1998normal}
I.~Das and J.~E. Dennis, ``Normal-boundary intersection: A new method for generating the {Pareto} surface in nonlinear multicriteria optimization problems,'' \emph{SIAM Journal on Optimization}, vol.~8, no.~3, pp. 631--657, 1998.

\bibitem{tuan2024framework}
T.~A. Tuan, L.~P. Hoang, D.~D. Le, and T.~N. Thang, ``A framework for controllable {Pareto} front learning with completed scalarization functions and its applications,'' \emph{Neural Networks}, vol. 169, pp. 257--273, 2023.

\bibitem{lin2022pareto2}
X.~Lin, Z.~Yang, X.~Zhang, and Q.~Zhang, ``{Pareto} set learning for expensive multi-objective optimization,'' \emph{In Proceedings of Advances in Neural Information Processing Systems}, vol.~35, pp. 19\,231--19\,247, 2022.

\bibitem{ruchte2021scalable}
M.~Ruchte and J.~Grabocka, ``Scalable {Pareto} front approximation for deep multi-objective learning,'' \emph{In Proceedings of IEEE International Conference on Data Mining (ICDM)}, pp. 1306--1311, 2021.

\bibitem{paszke2019pytorch}
A.~Paszke, S.~Gross, F.~Massa, A.~Lerer, J.~Bradbury, G.~Chanan, T.~Killeen, Z.~Lin, N.~Gimelshein, L.~Antiga \emph{et~al.}, ``Pytorch: An imperative style, high-performance deep learning library,'' \emph{In Proceedings of Advances in Neural Information Processing Systems}, vol.~32, 2019.

\bibitem{kingma2014adam}
D.~P. Kingma and J.~Ba, ``Adam: A method for stochastic optimization,'' \emph{arXiv preprint arXiv:1412.6980}, 2014.

\bibitem{zitzler2000comparison}
E.~Zitzler, K.~Deb, and L.~Thiele, ``Comparison of multiobjective evolutionary algorithms: Empirical results,'' \emph{Evolutionary Computation}, vol.~8, pp. 173--195, 2000.

\bibitem{deb2005scalable}
K.~Deb, L.~Thiele, M.~Laumanns, and E.~Zitzler, ``Scalable test problems for evolutionary multiobjective optimization,'' in \emph{Evolutionary Multiobjective Optimization: Theoretical Advances and Applications}.\hskip 1em plus 0.5em minus 0.4em\relax Springer, 2005, pp. 105--145.

\end{thebibliography}
\end{document}